\pdfoutput=1

\documentclass[11pt]{article}
\usepackage{EMNLP2023}

\usepackage{ragged2e}
\usepackage{tabularx}
\newcolumntype{Y}{>{\RaggedRight\arraybackslash}X} 
\newcolumntype{W}{>{\RaggedLeft\arraybackslash}X} 
\newcolumntype{H}{>{\centering\arraybackslash}X} 
\newcolumntype{Z}{>{\arraybackslash}X} 

\usepackage{times}
\usepackage{latexsym}
\usepackage{graphicx}
\usepackage[export]{adjustbox}
\usepackage[T1]{fontenc}
\usepackage[export]{adjustbox}[2011/08/13]
\usepackage{booktabs}
\usepackage{multirow}
\usepackage[utf8]{inputenc}

\usepackage{microtype}

\usepackage{inconsolata}
\usepackage{amsmath}
\usepackage{amsfonts}

\title{Clinical Text Deduplication Practices for Efficient Pretraining and Improved Clinical Tasks}

\author{Isotta Landi \\ CBIPM \\ Ichan School of Medicine \\ at Mount Sinai \\ New York \\ \texttt{isotta.landi2@mssm.edu} 
\And
  Eugenia Alleva \\ HPIMS \\ Icahn School of Medicine \\ at Mount Sinai \\ New York \And
  Alissa A. Valentine \\ CBIPM \\ Icahn School of Medicine \\ at Mount Sinai \\ New York \\
  \AND
  Lauren A. Lepow \\ CBIPM \\ Icahn School of Medicine \\ at Mount Sinai \\ New York 
  \And
  Alexander W. Charney \\ CBIPM \\ Icahn School of Medicine \\ at Mount Sinai \\ New York}

\begin{document}
\maketitle
\begin{abstract}
Despite being a unique source of information on patients' status and disease progression, clinical notes are characterized by high levels of duplication and information redundancy. In general domain text, it has been shown that deduplication does not harm language model (LM) pretraining, thus helping reduce the training cost. Although large LMs have proven to learn medical knowledge, they still require specialized domain adaptation for improved downstream clinical tasks. By leveraging large real-world clinical corpora, we first provided a fine-grained characterization of duplicates stemming from common writing practices and clinical relevancy. Second, we demonstrated that deduplicating clinical text can help clinical LMs encode less redundant information in a more efficient manner and do not harm classification tasks via prompt-based learning.
\end{abstract}

\section{Introduction}
Including clinical findings, decisions, and the deductive process of medicine, clinical notes are the most fine-grained documentation of a patient's interaction with a healthcare system \cite{percha2021}. As such, clinical notes present a unique opportunity, for the natural language processing (NLP) field, for the extraction of patient-level temporal information, such as patients' status and disease progression. 

The advent of the attention mechanism \cite{bahdanau2015} and the Transformer architecture \cite{vaswani2017} has enabled language models (LMs) to learn general-purpose features from text during pretraining and solve downstream tasks. In particular, large language models (LLMs), with billions of parameters, seem to already encode a great amount of medical knowledge. Nevertheless, because of medical jargon and abbreviations, clinical text is substantially different from both the general domain and the biomedical one. Moreover, compared to the general domain, the clinical corpora are characterized by significantly higher levels of duplicated content. It has been estimated that $33\% - 54.2\%$ of clinical text is duplicated \cite{steinkamp2022}, compared to less than $20\%$ for general corpora \cite{lee-etal-2022-deduplicating}. Relatively small, specialized, LMs have been shown to perform better than general-domain LLMs \cite{lehman-etal-2022-learning} in the clinical space. Hence, there is still the need for domain adaptation on large clinical datasets for better performance \cite{lehman2023} through pretraining on billions of tokens at a high computational cost. 

Both clinical LMs \cite{alsentzer-etal-2019-publicly, jiang2023} and LLMs \cite{yang2022} have been adapted to clinical text, further improving their performance on a variety of clinical tasks (e.g., medical question answering, natural language inference, and operational predictive tasks). The effects of duplicated content on clinical NLP systems have, to some extent, already been documented \cite{cohen2013, johnson2016, adams2021, searle2021}. Nevertheless, given the importance of domain adaptation and its computational cost for real-world healthcare application of LMs, there is still the need for a thorough characterization and identification of clinical text duplicates and a more rigorous analysis of the impact of different sources of duplication during both pretraining and for downstream tasks. Given the nature of clinical text as a highly duplicated and redundant corpus, we introduce here a step towards the answer to the question: ``Do we really need that much?''.

In this work, we considered a large dataset of $\sim4M$ clinical notes, obtained by combining ICU notes extracted from the MIMIC-III Clinical Database \cite{johnson2016} and the Mount Sinai Data Warehouse (MSDW), a large repository of healthcare information from the Mount Sinai Health System in New York City. We first extracted three types of duplicates, for a fine-grained characterization of the clinical corpora, i.e. within-note, between-note, and not clinically relevant duplicates. We then deduplicated the datasets at different levels by removing within-note, not relevant, and between-note duplicates, respectively. By investigating the perplexity (PPL) of a clinical LM adapted to deduplicated text, we showed that agnostic deduplication allows clinical LMs to encode less redundant information ($16\%$ PPL decrease), while retaining the medical knowledge to efficiently solve a downstream classification task (up to $\sim 1.5\%$ $F_1$ score increase) respect to the baselines. Moreover, ad hoc removal of not relevant clinical information does not harm clinical pretrained models' performance ($0.15\%$ $F_1$ score increase) on the 2006 n2c2 smoking status challenge via prompt-based learning \cite{uzuner2008}.

\section{Related Work}
 Most of the work related to clinical text duplication has been focusing on content identification and its within-/between-patient distribution \cite{wrenn2010, zhang2011, steinkamp2022} on English text, with few exceptions \cite{dhont2015}. 
 
 The negative influence of text duplicates on clinical NLP systems has been investigated by \citet{cohen2013} simulating different levels of duplicated text to investigate the performance variation for text-mining applications (i.e., collocation identification and topic modelling). Duplication in clinical text has also been linked to information redundancy measuring generative LLMs information encoding efficiency via PPL \cite{searle2021}. Such redundancy has been shown to range from $0.12\%$ up to $98\%$ in the MIMIC-III Clinical Database notes \cite{johnson2016}. Content redundancy also interferes with summarization efforts for hospital-course documentation \cite{adams2021} and has fostered calls for action \cite{wang2020}.

The most comprehensive work in terms of duplication identification, removal, and investigation of the impact on generative LLMs was done, for general domain English text, by \citet{lee-etal-2022-deduplicating}. Authors showed that, removing duplicates from text (1) does not hurt PPL; (2) reduces the model’s training cost; and (3) decreases the amount of memorized training data it generates, addressing privacy concerns.

In line with the approach of \citet{lee-etal-2022-deduplicating}, the main contributions of this work are:
\begin{itemize}
\item The conceptualization of three sources of duplication in the clinical text: within-note, between-note clinically relevant, and between-note clinically not relevant;
\item A classifier that enables the removal of clinically not relevant duplicated text with a precision of $0.98$;
\item An intrinsic evaluation of the impact of deduplication practices on LMs;
\item An extrinsic evaluation of the impact of deduplication on a downstream classification task via prompt-based learning.
\end{itemize}

Results from this study contribute to the clinical NLP landscape by (1) decreasing the computational costs of LM pretraining through the removal of unnecessary clinical text; (2) identifying clinically not relevant text that can be safely discarded to improve downstream tasks.

\section{Clinical Corpora}
This work was conducted on three main datasets. See preprocessing steps in Appendix~\ref{preproc}.

\subsection{ICU Notes}
We considered critical care unit notes from the MIMIC-III v1.4 database \cite{johnson2016} and obtained $2,083,180$ notes from $46,146$ patients. Median number of notes per patient (min/max) was $21\ (1/1,420)$. The total number of words after preprocessing was $482,668,224$. MIMIC-III includes deidentified notes from the critical care units of the Beth Israel Deaconess Medical Center, collected between 2001 and 2012.

For a real-world characterization of text duplication practices, on September 2022, we queried MSDW and extracted all ICU notes by (1) matching care site name to \emph{intensive}, \emph{critical}, or \emph{icu}; and (2) dropping entries labelled as ``Telephone Encounter''. The dataset included $2,119,557$ notes from $27,457$ patients. Median (min/max) number of notes per patient was $29\ (1/6,561)$. The total number of words after preprocessing was $1,068,388,637$.

\subsection{n2c2 Notes}
\label{sec:n2c2notes}
The n2c2 NLP Research Data Sets repository\footnote{\url{https://portal.dbmi.hms.harvard.edu/projects/n2c2-nlp/}} includes fully deidentified notes from the Research Patient Data Registry at Partners Healthcare System annotated for $9$ NLP challenges/workshops. We removed notes used for different challenges and merged train/test splits for all available tasks. We obtained a corpus of $5,798$ notes that was used for external validation of the relevance identification model.

\subsection{Smoking Challenge}
The dataset for the smoking status identification task consists of a subset of n2c2 notes and includes $398$ and $104$ notes in training/test sets, respectively. Notes are annotated as one of $5$ classes, i.e., \emph{current smoker}, \emph{smoker}, \emph{past smoker}, \emph{non smoker}, \emph{unknown} \cite{uzuner2008}. For more balanced classes, we merged ``current smoker'' and ``smoker'' into one.

\section{Duplication in Clinical Text}
\label{sec:did}
\begin{figure}
\centering
  \includegraphics[width=0.65\textwidth,center]{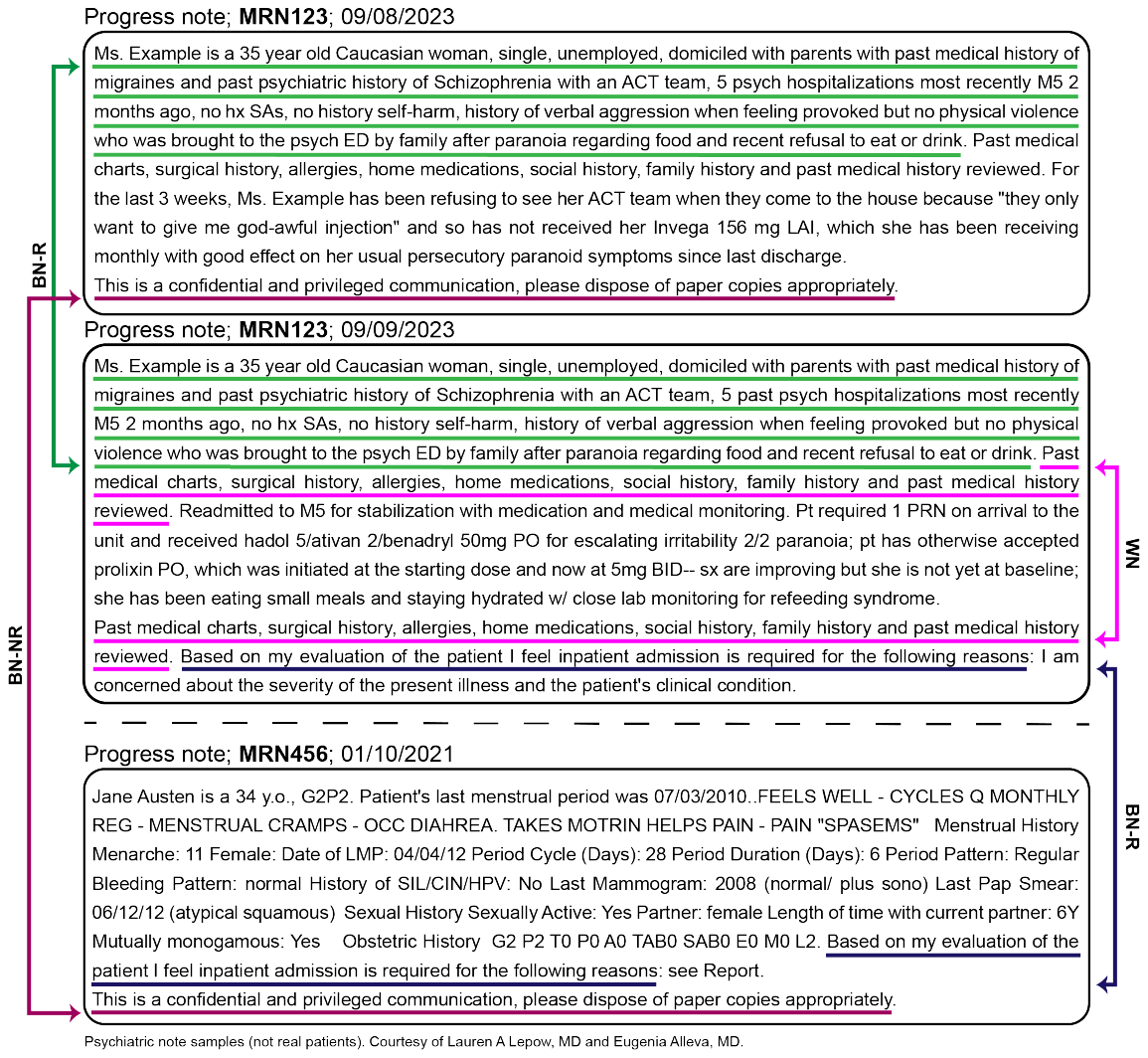}%
\caption{Examples of two progress reports for patient MRN123 and a progress report for patient MRN456 to showcase how duplicated text can be considered in clinical notes. In green, a copy-pasted portion of clinically relevant text between notes from the same patient (BN-R); in pink, a standardized sentence repeated within note (WN); in dark blue, a standardized relevant text repeated between notes from different patients (BN-R); and, in red, a portion of not clinically relevant text repeated between notes from different patients (BN-NR).}
 \label{fig:example}
\end{figure}

Duplicates in clinical text can arise from copy-paste practices, templates, and standardized text. The copy-paste practice consists in copying forward portions of a patient's previously written notes as a time-saving method to record past information that is still true at the time of writing or needs editing. Such duplication mainly occurs between notes from the same patient. On the other hand, templates and standardized text are sentences used as headers of required note sections or necessary for legal and billing purposes. As such, they can appear between notes from different patients, but also within a same note. Based on these observations, we distinguished duplicates in clinical text into two main groups: those occurring within a single note (WN) and those occurring between notes either from the same patient or from different patients (BN). In both cases, each duplicated sentence can either carry a clinically meaningful message or be not relevant as related to the patient's care and disease progression. For an example of how duplication can appear in clinical notes see Figure~\ref{fig:example}.

\subsection{Within-Note and Between-Note Duplicates}
To detect duplicated text, we processed MIMIC, MSDW, and n2c2 datasets separately. First, we identified sentences duplicated within a single note. Then, we applied the \textsc{ExactSubstr} algorithm developed by \citet{lee-etal-2022-deduplicating} to detect between-note duplicates. The algorithm leverages a suffix array to identify substrings that occur verbatim in more than one document. Because duplicates in clinical notes potentially span few sentences, we set the minimum matching substring length to $k=100$ and split the exact matches into sentences, only retaining those longer than $5$ characters. We extracted $\sim15M$ between-note duplicated sentences from MIMIC-III, $\sim28M$ from MSDW, and $207,007$ from the n2c2 dataset.

\subsection{Clinically Not Relevant Duplicates}
With the help of two trained physicians we further distinguished duplicates based on clinical relevance. Because, ideally, all within-note duplicates can be removed without further characterization, we defined not relevant repeated sentences as (1) repeated between notes; (2) not conveying any information on a patient's history or hospital stay; and (3) standard utterances used for legal purposes by the medical team. In the following, we will refer to repeated sentences that are not clinically relevant as BN-NR, whereas to repeated sentences that still convey a clinically relevant information as BN-R.  We defined seven topics that correspond to the description. Deidentified examples found in MIMIC-III and MSDW datasets, for each topic, are reported in Table~\ref{stab:1}.

\begin{table*}
\centering
\begin{tabularx}{\textwidth}{@{} Z | Z @{}}
\hline
\centering \textbf{MIMIC-III} & \multicolumn{1}{c}{\textbf{MSDW}}\\
\hline
\multicolumn{2}{c}{\textbf{Agreement with findings and care plan}}
\\\hline
\RaggedRight{I agree with his her note above including assessment and plan} & \RaggedRight{Attending physician note I have personally reviewed the images and resident interpretation thereof and agree with the findings}\\
\hline
\multicolumn{2}{c}{\textbf{Physical presence}}\\\hline
\RaggedRight{I saw and examined the patient and was physically present with the ICU Resident for key portions of the services provided} & \RaggedRight{I have fully participated in the care of this patient}\\
\hline
\multicolumn{2}{c}{\textbf{Discussion of the results (no consults)}}\\\hline
\RaggedRight{These findings were discussed with the referring physician} & \RaggedRight{This case was also discussed with Dr Jane Doe Dr John Doe}\\
\hline
\multicolumn{2}{c}{\textbf{Review of the results}}\\\hline
\RaggedRight{Results were personally reviewed with the MD caring for the patient} & \RaggedRight{I have reviewed the medical record discussed with house staff consulting services if any and nursing staff}\\
\hline
\multicolumn{2}{c}{\textbf{Notice}}\\\hline
\RaggedRight{Dr was notified in person of the results in the operating room at the time of the study} & \RaggedRight{}\\
\hline
\multicolumn{2}{c}{\textbf{Patient’s care}}\\\hline
\RaggedRight{The patient was monitored by a nurse throughout the procedure} & \RaggedRight{Patient presented and examined on rounds}\\
\hline
\multicolumn{2}{c}{\textbf{Information location (with no explicit clinical terms)}}\\\hline
\RaggedRight{See flowsheet for further details} & \RaggedRight{Attending note for full plan details by systems}\\
\hline
\multicolumn{2}{c}{\textbf{Hypothetical symptoms}}\\\hline
\RaggedRight{Please return to the hospital or call your PCP if you develop chest pain worsening shortness of breath increasing leg swelling dizziness or lightheadedness or if you have a new fever} & \RaggedRight{If you develop a fever chills redness or persistent discharge please contact the office during regular hours to report your symptoms }\\
\hline
\multicolumn{2}{c}{\textbf{Contact information}}\\\hline
\RaggedRight{Answering service will contact on call person during off hours Completed} & \RaggedRight{Please feel free to contact us if you should have any questions }\\
\hline
\multicolumn{2}{c}{\textbf{General information}}\\\hline
\RaggedRight{I would add the following remarks History nothing to add Physical Examination nothing to add Medical Decision Making nothing to add Total time spent on patient care 60 minutes Protected Section Addendum Entered NI 862 Name NI} & \RaggedRight{THIS IS A CONFIDENTIAL AND PRIVILEGED COMMUNICATION PLEASE DISPOSE OF PAPER COPIES APPROPRIATELY MSDW DISCHARGE SUMMARY REPORT PT NAME NAME}\\
\hline
\multicolumn{2}{c}{\textbf{Template procedural steps}}\\\hline
\RaggedRight{A preprocedural timeout and huddle was performed as per protocol} & \RaggedRight{All needle sponge and instrument counts were correct MD DD [DATE] TD [DATE] Job This is a privileged and confidential communication}\\
\hline
\end{tabularx}
\caption{Examples of duplicated not relevant information from MIMIC-III and MSDW.}
\label{stab:1}
\end{table*}

\paragraph{Task-adaptive Pretraining} To enable the automatic detection of not relevant duplicated sentences, we further pretrained the base version of the clinical masked language model GatorTron (i.e., $345$M parameters \citep{yang2022}) on $70\%$ of the combined MIMIC-III and MSDW unlabeled between-note duplicates (see Section~\ref{sec:did}) for task adaptation \cite{gururangan-etal-2020-dont}. The GatorTron model was downloaded from the NVIDIA NGC Catalog\footnote{\url{https://catalog.ngc.nvidia.com/orgs/nvidia/teams/clara/models/gatortron_og}}. We pretrained the model for $5$ epochs using a masked language modelling objective and learning rate $5e-5$. We scheduled linear warm-up ratio $0.01$ and weight decay $0.01$ within the AdamW optimizer \cite{loshchilov2018decoupled}. Maximum sequence length was $128$ and batch size was set at $64$.

\paragraph{Not Relevance Classification} The task-adapted model was fine-tuned for binary text classification to recognize duplicated sentences that are not clinically relevant. From the remaining set of between-note sentences ($\sim 8M$) that were not used for task adaptation, we randomly sampled $4,000$ sentences (equally for both clinical corpora) and manually annotated them as clinically ``relevant'' or ``not relevant''. At the end of this first run of annotations, we retained the same number of relevant/not relevant sentences that were split into train/dev/test sets. 

We fine-tuned the task-adapted GatorTron model on train/dev sets with different hyperparameter combinations and selected the model with highest precision on the dev set (see Appendix~\ref{sec:nrid}). The best model was evaluated on the test set through $F_1$ score, precision, and recall. We iterated the fine-tuning process until the model reached precision $>0.95$ on the test set. To prevent data leakage, at each iteration, we applied the model in inference on the unlabelled test set of duplicated sentences from the task-adaptation split and randomly extracted $2,000$ sentences from the ``not relevant'' class. We manually validated the new sentences and equally distributed them among the splits. The final number of annotated sentences per split can be found in Table~\ref{tab:1}, along with model's performance on the test set. Best model configuration was trained for $5$ epochs with batch size $8$, learning rate $2e-5$, weight decay $0.1$, and warm-up ratio $0.2$.

We externally validated the model by applying it to the between-note duplicated sentences identified in the n2c2 dataset. Then, we randomly extracted $100$ labelled sentences and manually validated them. We obtained a precision of $0.98$ and a recall of $1.0$.

\begin{table}
\centering
\begin{tabular}{lcc}
\hline
 & \textbf{MIMIC-III} & \textbf{MSDW}\\
\hline
Train (R/NR)  & $464/556$ & $455/573$\\
Dev (R/NR)  & $149/176$ & $131/186$\\
Test (R/NR)  & $141/176$ & $130/166$\\\hline
\multicolumn{1}{c}{$\mathbf{F_1}$} & \textbf{P} & \textbf{R}\\
\hline
\multicolumn{1}{c}{$0.84$} & $0.97$ & $0.80$\\\hline
\end{tabular}
\caption{Number of manually annotated relevant (R) and not relevant (NR) between-note duplicates for MSDW and MIMIC-III corpora. The performance of the task-adapted GatorTron model fine-tuned on the not relevant classification task is reported via $F_1$ score, precision (P), and recall (R) on the test set.}
\label{tab:1}
\end{table}

\paragraph{Deduplication Configurations}
 All between-note duplicates from MIMIC, MSDW, and n2c2 datasets were classified in inference to extract two subclasses from the between-note groups based on clinical relevance. To enable the investigation of the impact of different levels of duplication on clinical LMs and NLP tasks, we built $4$ deduplication configurations for each of the MIMIC-III, MSDW, and smoking identification datasets. 
 
 In the following, we will refer to the original dataset configuration as NONE; the dataset obtained by removing within-note duplicates only as WN; the dataset where both within-note and between-note not relevant duplicates were removed as WNNR; and the dataset where all duplicates were removed as WNBN. Table~\ref{tab:tokens} shows the progressive decrease in the number of words for the combined MIMIC-III and MSDW clinical corpora the less stringent the deduplication practice becomes. Duplicates are ubiquitous in both MIMIC-III and MSDW corpora and copy-forward information accounts for most of the clinically not relevant duplicates ($63-90\%$) and only $\sim 15\%$ of the relevant duplicates (see Appendix Table~\ref{tab:wp}). Further details on the distribution of duplicated sentences for all datasets can be found in Appendix Table~\ref{tab:dupstats}.

\begin{table}
\centering
\begin{tabular}{lccc}
\hline
&\textbf{Tokens}  & \textbf{\% Decrease} & \textbf{Words}\\\hline
NONE&$1.90$B &- &$1.48$B\\
WN&$1.89$B&$0.52$&$1.47$B\\
WNNR&$1.87$B&$1.57$&$1.45$B\\
WNBN&$1.12$B&$41.05$&$878$M\\\hline
\end{tabular}
\caption{Number of words and tokens for the combined MIMIC-III and MSDW deduplicated configurations. Percentage decrease in the number of tokens compared to the original datasets is reported for more stringent (i.e., WN/WNNR) and less stringent (i.e., WNBN) deduplication practices.}
\label{tab:tokens}
\end{table}

\section{Deduplication Impact on Clinical LMs}
\label{sec:dapt}
Similar to what was done by \citet{lazaridou2021} to assess the temporal generalization of LMs, we measured the impact of different deduplication strategies on the GatorTron model and its state-of-the-art counterpart ClinicalBERT \cite{alsentzer-etal-2019-publicly} via relative PPL. Such measure is more indicative of the learning process because, due to deduplication, some documents can be shorter than others between configurations and this can lower the PPL. Comparing the PPL obtained via further pretraining on deduplicated data to that measured via further pretraining on the non-deduplicated dataset we define relative PPL as:
\begin{equation*}
\frac{PPL\{M_{X^{tr}_{i}}(X^{dev}_{j})\} - PPL\{M_{X^{tr}_{NONE}}(X^{dev}_{j})\}}{PPL\{M_{X^{tr}_{NONE}}(X^{dev}_{j})\}}
\end{equation*}
with $i=\text{WN, WNNR, WNBN}$ and $j=\text{NONE, WN, WNNR, WNBN}$. $M_{X^{tr}_{i}}(X^{dev}_{j})\ \forall i,j$ indicates the model pretrained on $X^{tr}_{i}$ and evaluated on $X^{dev}_{j}$.

The masked language modelling objective makes encoder-based LMs, such as GatorTron, not suitable for computing PPL, because predictions are based on bi-directional context. Nevertheless, open-source generative models pretrained on clinical corpora are lacking, due to privacy concerns, and pretraining a LM from scratch comes with a high environmental cost. By masking the last token of each instance we computed PPL as: 
\begin{equation*}
PPL=\exp\{-\frac{1}{S}\sum_{i=1}^S\log\mathbb{P}(w^i_{k}|w^i_1, \cdots,w^i_{k-1})\}
\end{equation*} 
where $S$ is the total number of input instances and $w^i_j$ is the token in position $j$ for sentence $i$.

\paragraph{Deduplication Adaptation} For each deduplication configuration of combined MIMIC-III and MSDW, we randomly sampled $40,000$ notes that were evenly split into train/dev sets. We repeated the sampling $5$ times with different seeds. We further pretrained the GatorTron model on each training dataset for ``deduplication adaptation'' and computed PPL during inference on all other deduplicated dev sets. The PPL from each experiment was divided by the corresponding PPL of the model pretrained on NONE and evaluated on the same deduplication configuration. Further pretraining was done for $2,000$ steps with both masked language modelling and next sentence prediction objectives. Maximum sequence length was equal $128$, batch size $128$, and learning rate $1e-5$. In total, we pretrained $4\cdot 5$ models and performed $16\cdot 5$ evaluations.

\begin{figure}
\includegraphics[width=0.45\textwidth,left]{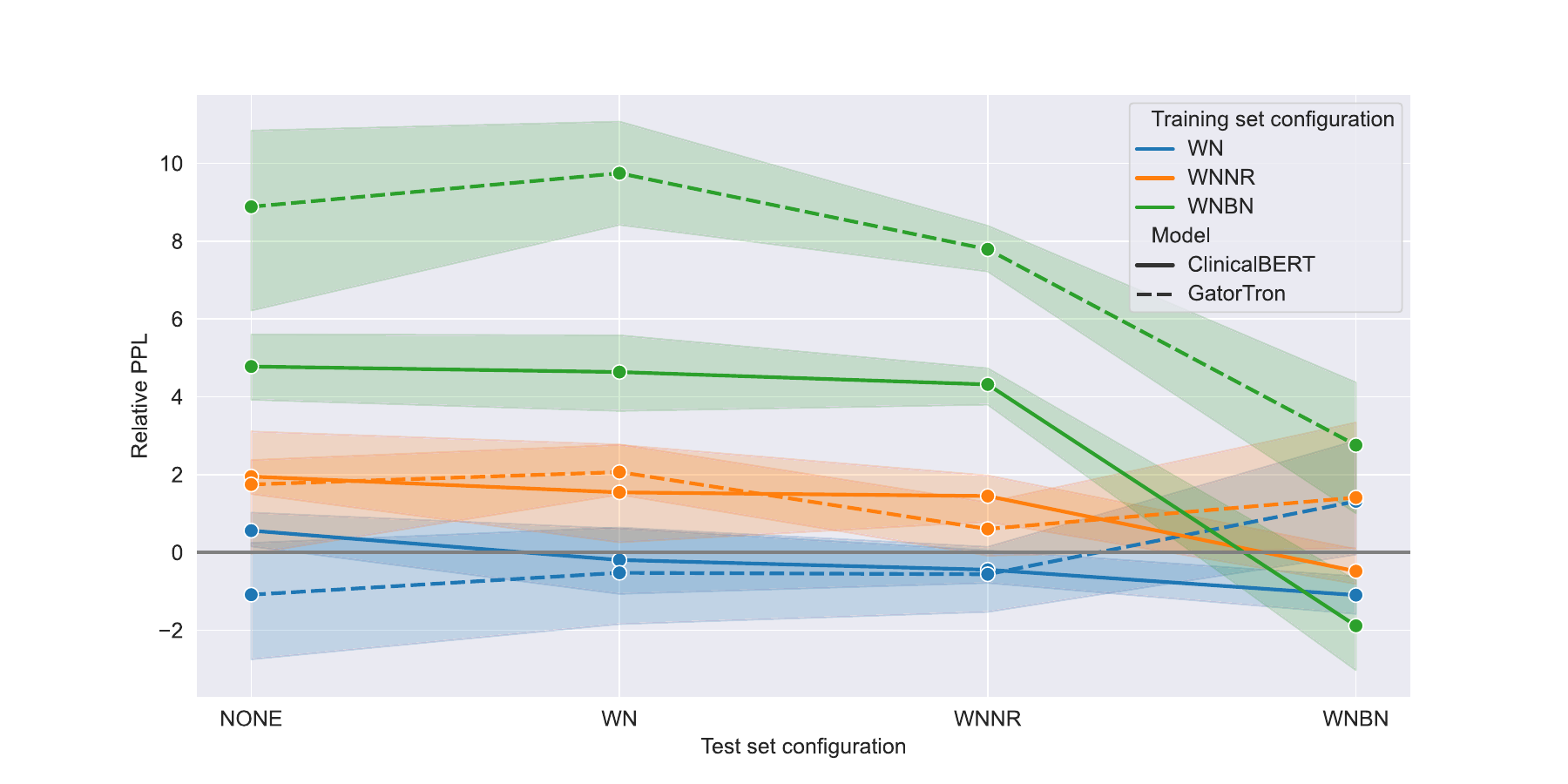}%
\caption{Mean relative PPL (sd) for LMs adapted to deduplicated clinical text.}
 \label{fig:1}
\end{figure}

\paragraph{PPL Investigation}
Figure~\ref{fig:1} shows the ability each pretrained ClinicalBERT- and GatorTron-based LMs has to generalize on deduplicated text relative to the baseline (i.e, models pretrained on NONE). If a line is close to $0$, it indicates that there is almost no difference in performance between the model (either ClinicalBERT or GatorTron) pretrained on NONE and the model pretrained on WN, WNNR, and WNBN datasets, when evaluated on all possible configurations. We can appreciate very similar behaviors for both ClinicalBERT- and GatorTron-based models. All trends for ClinicalBERT-based models are significantly decreasing towards WNBN, whereas only the GatorTron model pretrained on WNBN shows a significant decreasing trend ($p_s<0.001$ for t-tests on the slope of the fitted regression lines). Interestingly, ClinicalBERT pretrained on WN, WNNR, and WNBN, when evaluated on WNBN, reports a decrease in PPL respect to the baseline ($1.1\%$, $0.48\%$, and $1.9\%$ decrease, respectively). On the contrary, all GatorTron pretrained models show an increase in PPL, although GatorTron pretrained on both WN and WNNR has relative PPL close to zero. Both ClinicalBERT and GatorTron, when pretrained on WNBN and tested on either NONE, WN, and WNNR, report an increase in PPL of up to $9.7\%$ and $4.7\%$, respectively. Based on these results, we observe that agnostic deduplication (i.e., removal of all duplicates - WNBN) does not seem to impact the semantic and syntactic structure of the clinical text because relative PPL scores on WNBN are less than (maximum $1.88\%$ decrease for ClinicalBert) or close to (maximum $2.75\%$ average increase for GatorTron) zero for all pretrained models. If WNBN deduplication had disrupted the clinical language, we would have expected the model pretrained on WNBN to also show a poor generalization performance on WNBN itself. Moreover, from Table~\ref{tab:ppl}, we can observe that both LMs pretrained on WNBN and evaluated on NONE show a lower PPL score than the models pretrained on NONE and evaluated on WNBN. By computing the cross-entropy (i.e., $\log_2(PPL)$) we estimated the \emph{information content} for both clinical datasets, i.e., $5.14$ (for ClinicalBERT) and $5.16$ (for GatorTron) for WNBN and $4.88$ and $4.90$, respectively, for NONE. This suggests that LMs pretrained on clinical text with duplicates mostly encode redundant information, compared to models pretrained on deduplicated clinical notes. 

\begin{table}
\small
\centering
\begin{tabular}{llcc}
\hline
\textbf{Train} &       \textbf{Dev} &    \multicolumn{2}{c}{\textbf{PPL}} \\ 
\cline{3-4}
&&ClinicalBERT&GatorTron\\
\hline
   NONE &          NONE & $28.16\ (0.31)$ & $27.44\ (0.70)$ \\
   NONE &            WN$^*$ & $28.62\ (0.83)$ & $27.01\ (0.65)$ \\
   NONE &          WNNR & $27.74\ \ (0.70)$ & $26.79\ (0.62)$ \\
   NONE &          WNBN & $35.30\ (0.73)$ & $35.86\ (0.77)$ \\
   \hline\hline
   WN &          NONE & $28.32\ (0.29)$ & $27.13\ (0.39)$ \\
   WNNR &          NONE & $28.71\ (0.25)$ & $27.91\ (0.28)$ \\
   WNBN &          NONE$^{**}$ & $29.51\ (0.49)$ & $29.86\ (0.31)$ \\
   \hline
\end{tabular}
\caption{Mean (sd) PPL scores for ClinicalBERT and GatorTron models adapted on NONE and evaluated on all deduplication configurations and vice-versa. Pairwise t-tests were conducted for model comparison; $^*$ for $p<0.05$, $^{**}$ for $p<0.01$.}
\label{tab:ppl}
\end{table}

\section{Smoking Status Classification}
Last step was to investigate whether deduplicated datasets can lead to improved performance of clinical LMs on clinical NLP tasks, when used in pretraining or further tuning. The smoking status classification task was chosen because it is in line with the longstanding effort of the clinical NLP research community to extract salient information embedded in the clinical notes that is not present in structured electronic health records \cite{agrawal-etal-2022-large}.

The training set was split, with a $75/25$ ratio, into train/dev sets. We implemented a full-data learning approach with prompt \texttt{[Clinical note] smoking: [MASK]} and manual verbalizer \cite{schick-schutze-2021-exploiting}, leveraging the OpenPrompt library \cite{ding-etal-2022-openprompt}. The template was appended at the end of each note of the n2c2 smoking identification challenge dataset, with maximum sequence length of $512$ tokens. The best model was selected as the one with highest $F_1$ score on the dev set, when \texttt{[MASK]} was correctly predicted as either \emph{current smoker, past smoker, non-smoker}, or \emph{unknown}. The hyperparameter configurations used and the manual verbalizer implementation details can be found in Appendix~\ref{sec:nrid}. Best models' performance was assessed during evaluation on the test set through micro $F_1$ score, precision and recall. We repeated the entire learning paradigm (including model selection) $5$ times with different seeds. We ran the prompt-based learning paradigm on all deduplicated datasets for both the ClinicalBERT and GatorTron models and the corresponding further pretrained models from Section~\ref{sec:dapt}.

\paragraph{Deduplication Effects on Classification}
To investigate the effect of text deduplication on classification, we tuned ClinicalBERT and GatorTron models to classify patient's smoking status on all deduplication configurations. Table~\ref{tab:3} shows that ClinicalBERT's performance increases up to $14.18\%$ respect to the baseline, i.e., ClinicalBERT tuned on NONE, when tuned on WN and WNNR configurations. GatorTron, on the other hand, shows an increase in $F_1$ score of up to $3.85\%$. Both models, when tuned on WNBN, show a significant decrease ($p_s<0.05$, pairwise t-tests with Bonferroni correction) in $F_1$ scores of $11.90\%$ and $32.56\%$, respectively. The WNBN configuration includes notes $96\%$ shorter than the original smoking challenge dataset (see Appendix Table~\ref{tab:dupstats}). This suggests that most of the text describing patients' smoking status has probably been removed causing the drop in performance. The performance improvement when the model is tuned on the WN/WNNR configurations can be explained by the fact that those duplicates mostly occur at the end of each note. By design, they encompass copy-pasted text within notes and text written for legal and billing purposes. By removing those sentences the appended prompt appears to be learned more efficiently.

\paragraph{Deduplication Effects on Pretraining}
We then proceeded to apply the prompt-based paradigm to all models after adaptation on deduplicated datasets. By tuning them to solve the task on all combinations of the smoking challenge datasets, we aimed at assessing whether models pretrained on deduplicated text retain enough clinical knowledge to solve clinical tasks. Table~\ref{tab:3} shows scores for the top five results. Complete performances are shown in Figures~\ref{fig:smoking-cb},\ref{fig:smoking}, whereas $F_1$ scores by class can be found in Appendix Table~\ref{stab:3}. 

Both ClinicalBERT and GatorTron models, when pretrained on WNBN deduplicated clinical text, perform best if tuned on the original smoking status dataset (i.e., NONE), see Table~\ref{tab:3}. The performances are on-par with both the baseline models and the models pretrained and tuned on the original notes. 

The model adapted to the original dataset, when tuned on the smoking status dataset without not relevant sentences, shows an increase in performance compared to the baselines for ClinicalBERT ($19.65\%$ and $4.80\%$, respectively) and a decrease in performance of up to $0.92\%$ for GatorTron. Although the $F_1$ score comparisons not significant, we can conclude that pretrained models show a performance on WNNR configurations that is comparable to the standard pretraining/fine-tuning paradigm on datasets with retained not-relevant information.

We conclude that: (1) removing not relevant sentences does not harm LMs classification tasks; (2) the LMs pretrained on WNBN retains medical knowledge. This confirms that medical knowledge could be conveyed more efficiently through relatively smaller datasets with selected relevant information.

\begin{figure}
\centering
\includegraphics[width=0.45\textwidth,center]{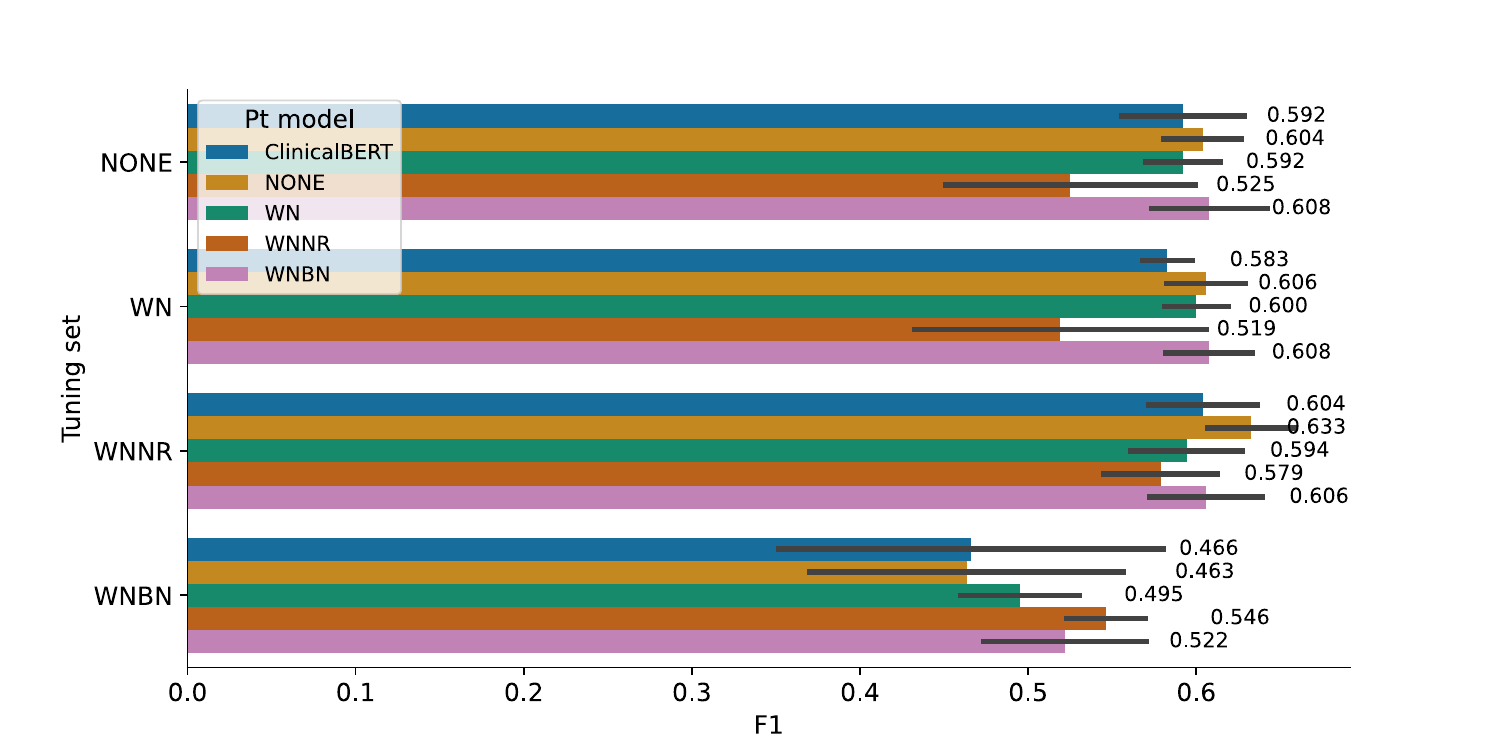}%
\caption{Mean $F_1$ scores (sd) for smoking status classification tasks with ClinicalBERT. Pretrained (Pt) models are tuned on all deduplication configurations of the n2c2 smoking dataset.}
 \label{fig:smoking-cb}
\end{figure}

\begin{figure}
\centering
\includegraphics[width=0.45\textwidth,center]{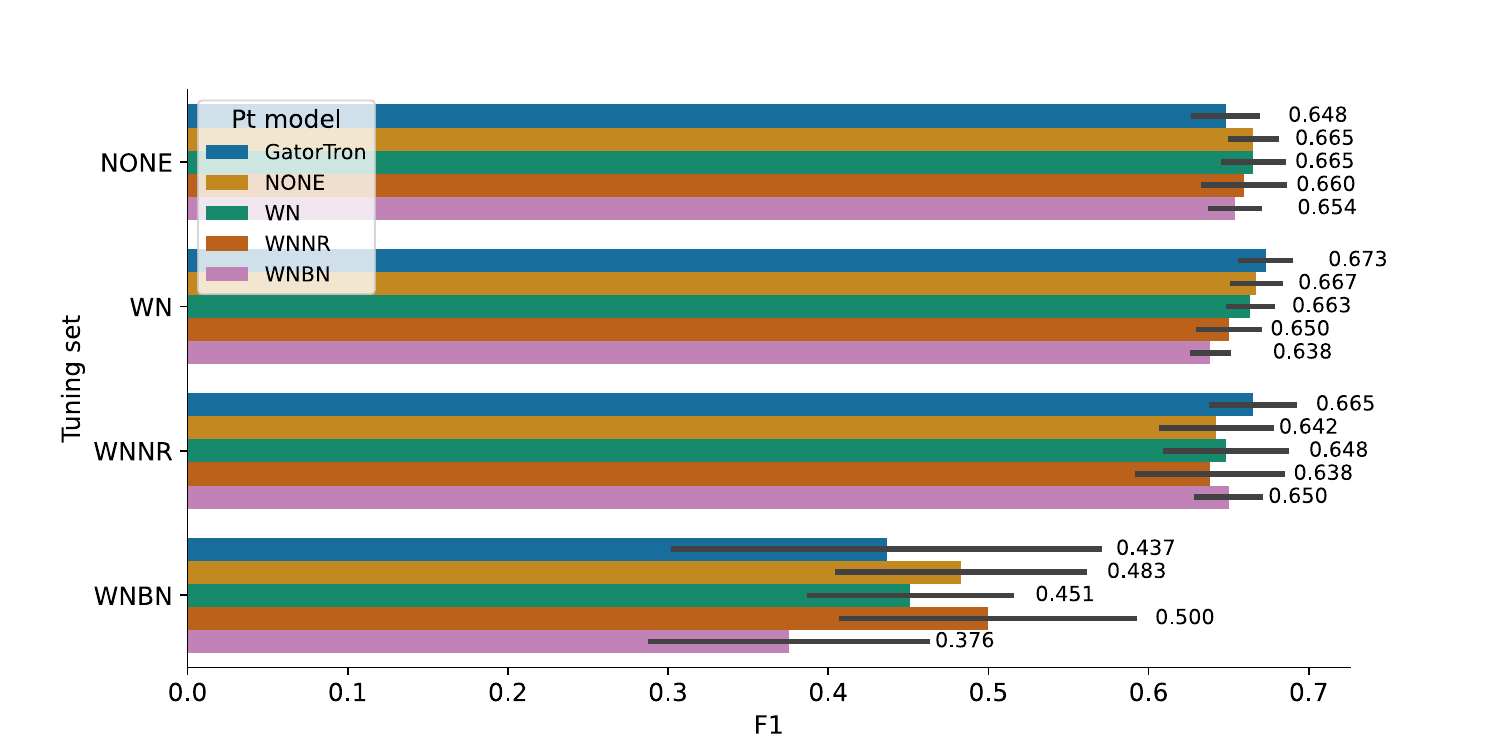}%
\caption{Mean $F_1$ scores (sd) for smoking status classification tasks with GatorTron. Pretrained (Pt) models are tuned on all deduplication configurations of the n2c2 smoking dataset.}
 \label{fig:smoking}
\end{figure}

\begin{table}
\small
\centering
\begin{tabular}{llc}        
\hline
\textbf{Pt model} & \textbf{Dataset} &  $\mathbf{F_1}$ \textbf{(sd)}\\\hline
\textbf{Baselines}&&\\\hline
ClinicalBERT & NONE & $0.529$ ($0.036$)\\
CB-NONE & NONE & $0.604$ ($0.023$)\\\hline
GatorTron & NONE &  $0.648$ ($0.02$)\\
GT-NONE & NONE & $0.665$ ($0.014$)\\\hline
\hline
\multicolumn{2}{l}{\textbf{Deduplication for tuning}}&\\\hline
ClinicalBERT & WN & $0.583$ ($0.015$)\\
ClinicalBERT & WNNR & $0.604$ ($0.032$)\\
ClinicalBERT & WNBN & $0.466$ ($0.114$)\\\hline
GatorTron & WN &  $\mathbf{0.673}$ ($0.015$)\\
GatorTron & WNNR &  $0.665$ ($0.026$)\\
GatorTron & WNBN &  $0.437$ ($0.133$)\\\hline
\hline
\multicolumn{2}{l}{\textbf{Deduplication for pretraining}}&\\\hline
CB-NONE & WNNR & $\textbf{0.633}$ ($0.026$)\\
CB-WN & WN & $0.6$ ($0.019$)\\
CB-WNNR & WNNR & $0.579$ ($0.034$)\\
CB-WNBN & NONE & $0.608$ ($0.034$)\\
CB-WNBN & WN & $0.608$ ($0.026$)\\\hline
GT-NONE & WN & $0.667$ ($0.015$)\\
GT-NONE & WNNR & $0.642$ ($0.034$)\\
GT-WN & NONE & $0.665$ ($0.018$)\\
GT-WNNR & NONE & $0.66$ ($0.025$)\\
GT-WNBN & NONE & $0.654$ ($0.015$)\\
GT-WNBN & WNNR & $0.65$ ($0.02$)\\\hline
\end{tabular}
\caption{Baselines: Model pretrained on non-deduplicated clinical text and tuned on non-deduplicated n2c2 smoking dataset. Deduplication for tuning: results for the ClinicalBERT/GatorTron models tuned on deduplicated datasets. Deduplication for pretraining: best results of deduplicated domain-adapted models tuned on all dataset configurations. CB: ClinicalBERT-based; GT: GatorTron-based.}
\label{tab:3}
\end{table}

\section{Discussion}
In this work, we have shown that clinical LMs can benefit from a fine-grained characterization of duplication within a clinical text. Leveraging a fine-tuned version of GatorTron for the classification of clinical relevance in sentences, we have shown that removing clinically not relevant information does not harm the performance of clinical language models on prompt-based classification tasks. Moreover, agnostic deduplication represents a promising alternative for clinical domain adaptation of general-purpose models on smaller datasets with selected medical information. This could lead to reduced computational cost and more efficient LMs information encodings that still retains medical knowledge for downstream tasks. 

The approach presented  can also be used for other downstream tasks that are heavily impacted by duplicates, e.g., summarization and temporal clinical information extraction. The clinical relevance detection could be expanded adding a temporal aspect at the patient level. This would enable the identification of copy-pasted utterances that, although referring to the patient's health history, record past states that are no longer relevant/true and should be ignored.

\section*{Limitations}
We acknowledge that this analysis was only done on clinical notes in the English language and does not necessarily generalize to other languages. Moreover, the conclusions on the advantages of pretraining a LM on agnostically deduplicated text are only limited to the further pretraining of already adapted clinical LMs. Although we did not perform a domain adaptation from scratch, the results can still benefit regular re-finetuning practices for data distribution shifts and hospital system changes. Although these results should be generalizable to other LMs, extending this approach to generative models, such as the GPT family of models \cite{radford2019}, would lead to a more comprehensive characterization of the impact of deduplication practices.

\section*{Ethics Statement}
This work uses deidentified documents from the publicly available n2c2 NLP Research Data Sets and MIMIC-III dataset. The use of the notes from MSDW dataset was approved by the hospital system IRB. The GatorTron model was downloaded from the NVIDIA NGC Catalog\footnote{\url{https://catalog.ngc.nvidia.com/orgs/nvidia/teams/clara/models/gatortron_og}} and used with few modifications with respect to the published version \cite{yang2022} to improve performance, as detailed in the paper. Code for clinical text deduplication and models tuning can be found at \url{https://github.com/landiisotta/clinical-duplicated-content}. 

\bibliography{emnlp2023}
\bibliographystyle{acl_natbib}

\appendix

\section{Preprocessing}
\label{preproc}
Each corpus was preprocessed with a custom sentencizer and tokenizer from the \texttt{en\_core\_sci\_md-0.5.1} \texttt{spacy v3.4.1} pipeline. A sentence was defined as the portion of text that ends with a punctuation mark or is an item in a list. Tokens were defined at the word level, with abbreviations, deidentified tokens, lab results, and medication dosages as one token. Dates and timestamps were replaced by special tokens \texttt{[DATE]} and \texttt{[TIME]}, respectively.

\section{Experiment Details}
\label{sec:nrid}
Experiments were implemented using the \texttt{huggingface}\footnote{\url{https://huggingface.co/}} library \cite{wolf-etal-2020-transformers} and Python $3.9$. All computations were performed on the high-performance computing resource of the MSDW health system using $1-4$ NVIDIA A100 GPUs (48Gb). Task- and domain-adaptation runs lasted $\sim40 - 48$ hours, whereas each run of all classification tasks lasted $10$ to $100$ seconds depending on the hyperparameter configuration. 

\paragraph{Not Relevance Classification} We tried different hyperparameter combinations for model selection. Batch size $\{8, 16, 32\}$; epochs $\{5, 10, 20\}$; learning rate $\{5e-6, 1e-5, 2e-5, 5e-5\}$; weight decay $\{0, 0.1, 0.01\}$: and warm-up ratio $\{0.01, 0.1, 0.2\}$.
\paragraph{Smoking Status Classification}
The hyperparameters for model selection were: batch size $\{2, 4, 8, 16\}$; learning rate $\{1e-5,2e-5,5e-5\}$; and steps $\{10,20,50,100\}$. The manual verbalizer was implemented has reported in Table~\ref{stab:verb}.

\begin{table}[h!]
\centering
\begin{tabular}{ll}
\hline
\textbf{Label} & \textbf{Words}\\\hline
 Non-Smoker & no, never\\
 Smoker & current, smoker\\
 Past Smoker & quit, stop\\
 Unknown & ``?'', ``.''\\ 
\hline
\end{tabular}
\caption{Manual verbalizer mapping for the smoking identification task.}
\label{stab:verb}
\end{table}



\begin{table*}
\small
\centering
\begin{tabular}{llccccc}
\hline
    &&&\textbf{Non-smoker} & \textbf{Smoker}& \textbf{Past smoker} & \textbf{Unknown}\\
    \cline{4-7}
    \textbf{Pt model} & \textbf{Dataset} &        \textbf{$F_1$ (sd)} &        \textbf{$F_1$ (sd)} &        \textbf{$F_1$ (sd)} &        \textbf{$F_1$ (sd)} &        \textbf{$F_1$ (sd)} \\
\hline
ClinicalBERT &    NONE & $0.592$ ($0.036$) & $0.308$ ($0.118$) & $0.236$ ($0.121$) &   $0.1$ ($0.141$) &  $0.775$ ($0.03$) \\
ClinicalBERT &      WN & $0.583$ ($0.015$) & $0.304$ ($0.038$) & $0.223$ ($0.053$) &  $0.162$ ($0.15$) & $0.774$ ($0.016$) \\
ClinicalBERT &    WNNR & $\mathbf{0.604}$ ($0.032$) & $0.361$ ($0.097$) & $0.309$ ($0.033$) & $0.135$ ($0.097$) &  $0.774$ ($0.02$) \\
ClinicalBERT &    WNBN & $0.466$ ($0.114$) & $0.127$ ($0.145$) &  $0.156$ ($0.15$) &     $0.0$ ($0.0$) &  $0.63$ ($0.113$) \\
\hline
     CB-NONE &    NONE & $0.604$ ($0.023$) & $0.336$ ($0.038$) & $0.294$ ($0.066$) & $0.211$ ($0.185$) & $0.768$ ($0.013$) \\
     CB-NONE &      WN & $0.606$ ($0.024$) & $0.332$ ($0.057$) & $0.292$ ($0.054$) & $0.194$ ($0.144$) & $0.771$ ($0.011$) \\
     CB-NONE &    WNNR & $\mathbf{0.633}$ ($0.026$) & $0.407$ ($0.043$) & $0.265$ ($0.075$) & $0.276$ ($0.173$) &  $0.786$ ($0.01$) \\
     CB-NONE &    WNBN & $0.463$ ($0.093$) & $0.142$ ($0.103$) & $0.209$ ($0.128$) &  $0.027$ ($0.06$) & $0.624$ ($0.095$) \\
     \hline
       CB-WN &    NONE & $0.592$ ($0.022$) & $0.307$ ($0.089$) & $0.219$ ($0.058$) & $0.093$ ($0.091$) & $0.772$ ($0.018$) \\
       CB-WN &      WN &   $\mathbf{0.6}$ ($0.019$) & $0.329$ ($0.065$) & $0.292$ ($0.071$) &   $0.2$ ($0.144$) & $0.761$ ($0.014$) \\
       CB-WN &    WNNR & $0.594$ ($0.033$) & $0.294$ ($0.076$) & $0.351$ ($0.033$) & $0.123$ ($0.149$) & $0.776$ ($0.025$) \\
       CB-WN &    WNBN & $0.495$ ($0.035$) & $0.093$ ($0.053$) & $0.199$ ($0.122$) &     $0.0$ ($0.0$) & $0.678$ ($0.038$) \\
       \hline
     CB-WNNR &    NONE & $0.525$ ($0.074$) & $0.201$ ($0.084$) & $0.036$ ($0.081$) & $0.018$ ($0.041$) & $0.699$ ($0.075$) \\
     CB-WNNR &      WN & $0.519$ ($0.087$) & $0.206$ ($0.109$) &  $0.06$ ($0.134$) &  $0.02$ ($0.045$) & $0.683$ ($0.093$) \\
     CB-WNNR &    WNNR & $\mathbf{0.579}$ ($0.034$) & $0.165$ ($0.066$) &     $0.0$ ($0.0$) &     $0.0$ ($0.0$) &  $0.741$ ($0.02$) \\
     CB-WNNR &    WNBN & $0.546$ ($0.023$) &     $0.0$ ($0.0$) & $0.097$ ($0.069$) &     $0.0$ ($0.0$) &  $0.72$ ($0.014$) \\
     \hline
     CB-WNBN &    NONE & $\mathbf{0.608}$ ($0.034$) & $0.346$ ($0.156$) &  $0.32$ ($0.079$) & $0.211$ ($0.054$) & $0.774$ ($0.018$) \\
     CB-WNBN &      WN & $\mathbf{0.608}$ ($0.026$) & $0.311$ ($0.091$) & $0.301$ ($0.072$) & $0.268$ ($0.217$) & $0.769$ ($0.011$) \\
     CB-WNBN &    WNNR & $0.606$ ($0.033$) & $0.318$ ($0.069$) & $0.304$ ($0.061$) & $0.159$ ($0.126$) & $0.774$ ($0.019$) \\
     CB-WNBN &    WNBN & $0.522$ ($0.048$) &   $0.2$ ($0.152$) & $0.077$ ($0.113$) &     $0.0$ ($0.0$) & $0.694$ ($0.039$) \\
\hline \hline
   GatorTron &    NONE &  $0.648$ ($0.02$) & $0.553$ ($0.028$) & $0.254$ ($0.109$) & $0.331$ ($0.046$) & $0.794$ ($0.013$) \\
   GatorTron &      WN & $\mathbf{0.673}$ ($0.015$) & $0.564$ ($0.058$) &  $0.34$ ($0.055$) & $0.377$ ($0.081$) & $0.807$ ($0.011$) \\
   GatorTron &    WNNR & $0.665$ ($0.026$) & $0.544$ ($0.042$) & $0.358$ ($0.067$) & $0.336$ ($0.167$) & $0.804$ ($0.008$) \\
   GatorTron &    WNBN & $0.437$ ($0.133$) &  $0.19$ ($0.051$) & $0.185$ ($0.081$) &     $0.0$ ($0.0$) & $0.594$ ($0.144$) \\
\hline
     GT-NONE &    NONE & $0.665$ ($0.014$) & $0.557$ ($0.037$) & $0.347$ ($0.077$) & $0.331$ ($0.061$) & $0.802$ ($0.013$) \\
     GT-NONE &      WN & $\mathbf{0.667}$ ($0.015$) & $0.549$ ($0.057$) &  $0.313$ ($0.08$) & $0.389$ ($0.058$) & $0.803$ ($0.007$) \\
     GT-NONE &    WNNR & $0.642$ ($0.034$) & $0.526$ ($0.057$) & $0.298$ ($0.081$) & $0.296$ ($0.076$) & $0.791$ ($0.029$) \\
     GT-NONE &    WNBN & $0.483$ ($0.077$) &  $0.265$ ($0.08$) & $0.315$ ($0.078$) & $0.036$ ($0.081$) & $0.628$ ($0.091$) \\
     \hline
       GT-WN &    NONE & $\mathbf{0.665}$ ($0.018$) & $0.547$ ($0.042$) &  $0.352$ ($0.04$) & $0.322$ ($0.114$) & $0.806$ ($0.004$) \\
       GT-WN &      WN & $0.663$ ($0.014$) & $0.544$ ($0.027$) & $0.347$ ($0.044$) &   $0.325$ ($0.1$) &  $0.804$ ($0.01$) \\
       GT-WN &    WNNR & $0.648$ ($0.038$) &  $0.529$ ($0.04$) & $0.331$ ($0.028$) & $0.361$ ($0.102$) & $0.784$ ($0.032$) \\
       GT-WN &    WNBN & $0.451$ ($0.063$) & $0.143$ ($0.108$) & $0.289$ ($0.036$) & $0.024$ ($0.053$) & $0.613$ ($0.066$) \\
       \hline
     GT-WNNR &    NONE &  $\mathbf{0.66}$ ($0.025$) & $0.546$ ($0.029$) & $0.319$ ($0.106$) &  $0.323$ ($0.07$) & $0.802$ ($0.011$) \\
     GT-WNNR &      WN &  $0.65$ ($0.019$) & $0.534$ ($0.043$) & $0.325$ ($0.058$) & $0.268$ ($0.088$) &  $0.796$ ($0.01$) \\
     GT-WNNR &    WNNR & $0.638$ ($0.045$) &  $0.525$ ($0.04$) & $0.348$ ($0.028$) &  $0.318$ ($0.11$) & $0.776$ ($0.046$) \\
     GT-WNNR &    WNBN &   $0.5$ ($0.091$) & $0.164$ ($0.103$) & $0.159$ ($0.147$) &     $0.0$ ($0.0$) & $0.659$ ($0.099$) \\
     \hline
     GT-WNBN &    NONE & $\mathbf{0.654}$ ($0.015$) & $0.538$ ($0.043$) & $0.328$ ($0.019$) & $0.249$ ($0.083$) & $0.801$ ($0.012$) \\
     GT-WNBN &      WN & $0.638$ ($0.011$) & $0.539$ ($0.043$) & $0.294$ ($0.037$) &  $0.26$ ($0.046$) & $0.788$ ($0.017$) \\
     GT-WNBN &    WNNR &   $0.65$ ($0.02$) & $0.541$ ($0.007$) &  $0.308$ ($0.02$) & $0.321$ ($0.012$) & $0.794$ ($0.017$) \\
     GT-WNBN &    WNBN & $0.376$ ($0.086$) & $0.226$ ($0.111$) &  $0.34$ ($0.082$) & $0.036$ ($0.081$) &  $0.48$ ($0.124$) \\
\hline
\end{tabular}
\caption{Mean $F_1$ scores (sd) overall and by class for the smoking status identification tasks. CB: ClinicalBERT-based; GT: GatorTron-based.}
\label{stab:3}
\end{table*}

\begin{table*}
\small
\centering
\begin{tabularx}{\textwidth}{@{}W H H H H @{}}
\hline
 \textbf{Deduplication Configurations}& \textbf{\% Notes with at least one duplicate} & \textbf{Median duplicated sentences per note (min/max)} & \textbf{Median number of words per note (min/max)} & \textbf{Median number of words reduction ($\%$)}\\
\hline
MIMIC-NONE & - & - & $139\ (1/7,714)$\\
MIMIC-WN & $5.39$ & $1\ (1/326)$ & $139\ (1/7,522)$ & $0.00$\\
MIMIC-WNNR & $8.81$ & $1\ (1/25)$ & $123\ (1/7508)$ & $11.51$\\
MIMIC-WNBN & $53.11$ & $6\ (1/310)$ & $10\ (1/6,053)$ & $92.80$\\
\hline\hline
MSDW-NONE & - & - & $277\ (1/18,579)$\\
MSDW-WN & $8.65$ & $2\ (1/593)$ & $141\ (1/13066)$ & $49.09$\\
MSDW-WNNR & $29.28$ & $1\ (1/35)$ & $126\ (1.0/12,814)$ & $54.51$\\
MSDW-WNBN & $74.49$ & $8\ (1/704)$ & $9\ (0/9,312)$ & $96.75$\\\hline
\hline
smoking-NONE & - & - & $634\ (72/2,752)$\\
smoking-WN & $8.96$ & $38\ (1/193)$ & $629.5\ (72/2,752)$ & $0.71$\\
smoking-WNNR & $80.27$ & $38\ (1/193)$ & $604.5\ (56/2,712$ & $4.65$\\
smoking-WNBN & $97.96$ & $37.5\ (1/192)$ & $20\ (2/1,423)$ & $96.84$\\\hline
\end{tabularx}
\caption{Duplicated sentences distribution and percentage reduction of note length for MIMIC-III, MSDW, and smoking task dataset deduplication configurations.}
\label{tab:dupstats}
\end{table*}

\begin{table*}
\small
\centering
\begin{tabularx}{\textwidth}{@{} H Z H @{}}
\hline
\centering \textbf{Duplicates} & \centering \textbf{Note types} & \textbf{\RaggedLeft{Percentage of copy-forward sentences}}\\\hline
MIMIC-WN & \RaggedRight{ECG} & -\\
MSDW-WN & \RaggedRight{L\&D delivery note} & -\\\hline\hline
MIMIC-NR & \RaggedRight{General} & $63.89\ (0/100)$\\
MSDW-NR & \RaggedRight{OR preop anesthesia, Miscellaneous, Observation attestation, Pharmacist notes} & $90.48\ (0/100)$ \\\hline\hline
MSDW-BN & \RaggedRight{Nursing, Physician} & $15.46\ (0/100)$\\
MSDW-BN & \RaggedRight{Miscellaneous, OR postop, H\&P, ED disposition decision, Observation attestation, Interval H\&P note} & $11.38\ (0/100)$\\\hline
\end{tabularx}
\caption{Note types with highest percentage of duplicated sentences. Within-patient median percentage (min/max) of carry-forward sentences.}
\label{tab:wp}
\end{table*}

\end{document}